\documentclass[10pt,twocolumn,letterpaper]{article}

\usepackage{iccv}
\usepackage{times}
\usepackage{epsfig}
\usepackage{graphicx}
\usepackage{amsmath}
\usepackage{amssymb}
\usepackage{booktabs} 
\usepackage{color}
\usepackage{cite} 
\usepackage{authblk} 
\usepackage{balance}

\definecolor{julieta_color}{RGB}{56,108,176} 
\definecolor{javier_color}{RGB}{217,95,2}    
\definecolor{rayat_color}{RGB}{228,26,28}   

\usepackage[pagebackref=true,breaklinks=true,letterpaper=true,colorlinks,bookmarks=false]{hyperref}

\iccvfinalcopy 

\def\iccvPaperID{****} 

\ificcvfinal\fi
\begin{document}

\title{A simple yet effective baseline for 3d human pose estimation\vspace{-1ex}} 

\author[1]{Julieta Martinez}
\author[1]{Rayat Hossain}
\author[2]{Javier Romero}
\author[1]{James J. Little}
\affil[1]{University of British Columbia, Vancouver, Canada}
\affil[2]{Body Labs Inc., New York, NY\vspace{0.5ex}}
\affil[ ]{\tt\small julm@cs.ubc.ca, rayat137@cs.ubc.ca, javier.romero@bodylabs.com, little@cs.ubc.ca\vspace{-2ex}}
\maketitle

\begin{abstract}
Following the success of deep convolutional networks, state-of-the-art methods for 3d human pose estimation have focused on deep end-to-end systems that predict 3d joint locations given raw image pixels.
Despite their excellent performance, it is often not easy to understand whether their remaining error stems from a limited 2d pose (visual) understanding, or from a failure to map 2d poses into 3-dimensional positions.

With the goal of understanding these sources of error, we set out to build a system that given 2d joint locations predicts 3d positions. Much to our surprise, we have found that, with current technology, ``lifting'' ground truth 2d joint locations to 3d space is a task that can be solved with a remarkably low error rate: a relatively simple deep feed-forward network outperforms the best reported result by about 30\% on Human3.6M, the largest publicly available 3d pose estimation benchmark. Furthermore, training our system on the output of an off-the-shelf state-of-the-art 2d detector (\ie, using images as input) yields state of the art results --
this includes an array of systems that have been trained end-to-end specifically for 
this task.
Our results indicate that a large portion of the error of 
modern
deep 3d pose estimation systems stems from their visual analysis, and suggests directions to further advance the state of the art in 3d human pose estimation.
\end{abstract}

\section{Introduction}

The vast majority of existing depictions of humans are two dimensional, \eg video footage, images or paintings. These representations have traditionally played an important role in conveying facts, ideas and feelings to other people, and this way of transmitting information has only been possible thanks to the ability of humans to understand complex spatial arrangements in the presence of depth ambiguities. For a large number of applications, including virtual and augmented reality, apparel size estimation or even autonomous driving, giving this spatial reasoning power to machines is crucial. In this paper, we will focus on a particular instance of this spatial reasoning problem: 3d human pose estimation from a single image.

More formally, given an image -- a 2-dimensional representation -- of a human being, 3d pose estimation is the task of producing a 3-dimensional figure that matches the spatial position of the depicted person.
In order to go from an image to a 3d pose, an algorithm has to be invariant to a number of factors, including background scenes, lighting, clothing shape and texture, skin color and image imperfections, among others.
Early methods achieved this invariance through features such as silhouettes~\cite{AgarwalT04}, shape context~\cite{bb93102}, SIFT descriptors~\cite{bb41073} or edge direction histograms~\cite{bb93266}.
While data-hungry deep learning systems currently outperform approaches based on human-engineered features on tasks such as 2d pose estimation (which also require these invariances), the lack of 3d ground truth posture data for images in the wild makes the task of inferring 3d poses directly from colour images challenging.

Recently, some systems have explored the possibility of directly inferring 3d poses from images with end-to-end deep architectures~\cite{tekin2016structured,volumetric}, and other systems argue that 3d reasoning from colour images can be achieved by training on synthetic data~\cite{RogezS16, varol_2017}. In this paper, we explore the power of decoupling 3d pose estimation into the well studied problems of 2d pose estimation~\cite{cpm, stacked-hourglass}, and 3d pose estimation from 2d joint detections, focusing on the latter.
Separating pose estimation into these two problems gives us the possibility of exploiting existing 2d pose estimation systems, which already provide invariance to the previously mentioned factors. Moreover, we can train data-hungry algorithms for the 2d-to-3d problem with large amounts of 3d mocap data captured in controlled environments, while working with low-dimensional representations that scale well with large amounts of data.

Our main contribution to this problem is the design and analysis of a neural network that performs slightly better than state-of-the-art systems (increasing its margin when the detections are fine-tuned, or ground truth) and is fast (a forward pass takes around $3$ms on a batch of size 64, allowing us to process as many as $300$ fps in batch mode), while being easy to understand and reproduce. The main reason for this leap in accuracy and performance is a set of simple ideas, such as estimating 3d joints in the camera coordinate frame, adding residual connections and using batch normalization. These ideas could be rapidly tested along with other unsuccessful ones (\eg estimating joint angles) due to the simplicity of the network.

The experiments show that inferring 3d joints from groundtruth 2d projections can be solved with a surprisingly low error rate -- 30\% lower than state of the art -- on the largest existing 3d pose dataset.
Furthermore, training our system on noisy outputs from a recent 2d keypoint detector yields results that 
slightly outperform
the state-of-the-art on 3d human pose estimation, which comes from systems trained end-to-end from raw pixels.

Our work considerably improves upon the previous best 2d-to-3d pose estimation result using noise-free 2d detections in Human3.6M, while also using a simpler architecture. This shows that lifting 2d poses is, although far from solved, an easier task than previously thought.
Since our work also achieves 
state-of-the-art results
starting from the output of an off-the-shelf 2d detector, it also suggests that current systems could be further improved by focusing on the visual parsing of human bodies in 2d images.
Moreover, we provide and release a high-performance, yet lightweight and easy-to-reproduce
baseline that sets a new bar for future work in this task.
Our code is publicly available at \url{https://github.com/una-dinosauria/3d-pose-baseline}.

\section{Previous work}

\paragraph{Depth from images}
The perception of depth from purely 2d stimuli is a classic problem that has captivated the attention of scientists and artists at least since the Renaissance, when Brunelleschi used the mathematical concept of perspective to convey a sense of space in his paintings of Florentine buildings.

Centuries later, similar perspective cues have been exploited in computer vision to infer lengths, areas and distance ratios in arbitrary scenes~\cite{zisserman99}. Apart from perspective information, 
classic computer vision systems have tried to use other cues like shading~\cite{zhang_shapefromshading} or  texture~\cite{lindeberg_shapefromtexture} to recover depth from a single image. Modern systems~\cite{saxena_2D_3D, fergus_2D_3D_iccv2015, liu_2d_3d_cvpr_2015, popa_cvpr17} typically approach this problem from a supervised learning perspective, letting the system infer which image features are most discriminative for depth estimation.

\paragraph{Top-down 3d reasoning}
One of the first algorithms for depth estimation took a different approach: exploiting the known 3d structure of the objects in the scene~\cite{roberts_63}. 
It has been shown that this top-down information is also used by humans when perceiving human motion abstracted into a set of sparse point projections~\cite{bulthoff1998top}. The idea of reasoning about 3d human posture from a minimal representation such as sparse 2d projections, abstracting away other potentially richer image cues, has inspired the problem of 3d pose estimation from 2d joints that we are 
addressing 
in this work.

\paragraph{2d to 3d joints}
The problem of inferring 3d joints from their 2d projections can be traced back to the classic work of Lee and Chen~\cite{Chen85b}. They showed that, given the bone lengths, the problem boils down to a binary decision tree where each split correspond to two possible states of a joint with respect to its parent.
This binary tree can be pruned based on joint constraints, though it rarely resulted in a single solution. Jiang~\cite{jiang_10} used a large database of poses to resolve ambiguities based on nearest neighbor queries. Interestingly, the idea of exploiting nearest neighbors for refining the result of pose inference has been recently revisited by Gupta~\etal~\cite{gupta20143dpose}, who incorporated temporal constraints during search, and by Chen and Ramanan~\cite{ChenR16a}. Another way of compiling knowledge about 3d human pose from datasets is by creating overcomplete bases suitable for representing human poses as sparse combinations~\cite{Ramakrishna:2012, wang2014robust,akhter-and-black,keep-it-simpl,zhou2016sparseness,zhou2016sparse}, 
lifting the pose to a reproducible kernel Hilbert space (RHKS)~\cite{ionescu_cvpr14}
or by creating novel priors from specialized datasets of extreme human poses~\cite{akhter-and-black}.

\begin{figure*}
  \includegraphics[width=\linewidth,trim=0mm 100mm 0mm 75mm,clip=true]{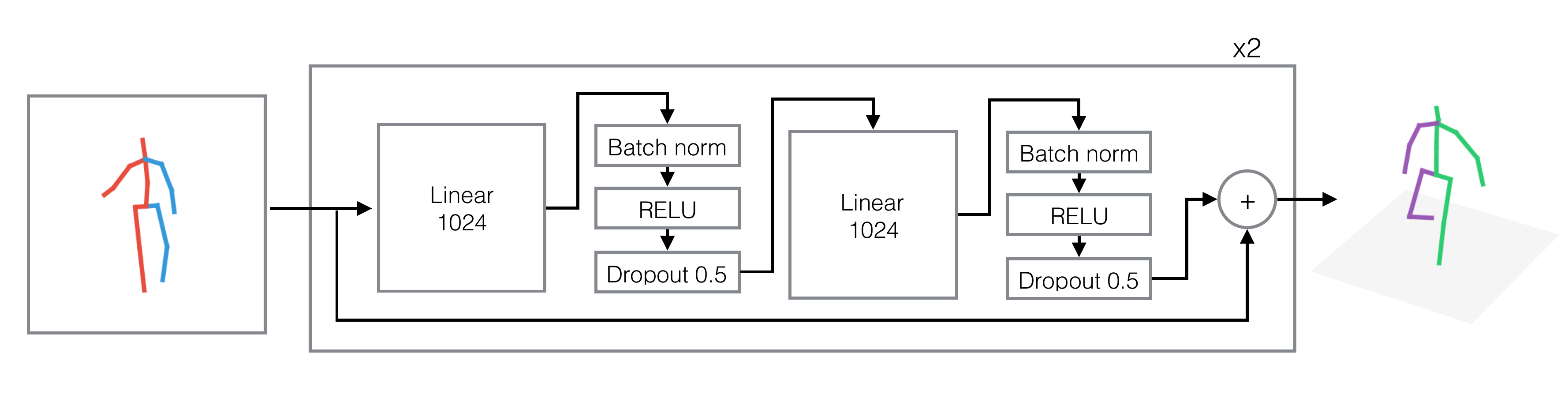}
  \vspace{-8mm}
  \caption{A diagram of our approach. The building block of our network is a linear layer, followed by batch normalization, dropout and a RELU activation. This is repeated twice, and the two blocks are wrapped in a residual connection. The outer block is repeated twice. The input to our system is an array of 2d joint positions, and the output is a series of joint positions in 3d.}
  \label{fig:diag}
\end{figure*}

\paragraph{Deep-net-based 2d to 3d joints}
Our system is most related to recent work that learns the mapping between 2d and 3d with deep neural networks.
Pavlakos~\etal~\cite{volumetric} introduced a deep convolutional neural network based on the stacked hourglass architecture~\cite{stacked-hourglass} that, instead of regressing 2d joint probability heatmaps, maps to probability distributions in 3d space.
Moreno-Noguer~\cite{distance-matrix} learns to predict a pairwise distance matrix (DM) from 2-to-3-dimensional space. Distance matrices are invariant up to rotation, translation and reflection; therefore, multi-dimensional scaling is complemented with a prior of human poses~\cite{akhter-and-black} to rule out unlikely predictions.

A major motivation behind 
Moreno-Noguer's
DM regression approach, as well as the volumetric approach of Pavlakos~\etal, is the idea that predicting 3d keypoints from 2d detections is inherently difficult. For example, Pavlakos~\etal~\cite{volumetric} present a baseline where a direct 3d joint representation (such as ours) is used instead (Table 1 in~\cite{volumetric}), with much less accurate results than using volumetric regression\footnote{This approach, however, is slightly different from ours, as the input is still image pixels, and the intermediate 2d body representation is a series of joint heatmaps -- not joint 2d locations.}
Our work contradicts the idea that regressing 3d keypoints from 2d joint detections directly should be avoided, and shows that a well-designed and simple network can perform quite competitively in the task of 2d-to-3d keypoint regression.

\paragraph{2d to 3d angular pose}
There is a second branch of algorithms for inferring 3d pose from images which estimate the body configuration in terms of angles (and sometimes body shape) instead of directly estimating the 3d position of the joints~\cite{barron_2001,ParameswaranC04,keep-it-simpl,zhou2016deep}. The main advantages of these methods are that the dimensionality of the problem is lower due to the constrained mobility of human joints, and that the resulting estimations are forced to have a human-like structure.
Moreover, constraining human properties such as bone lengths or joint angle
ranges is rather simple with this representation~\cite{WeiC09}.
We have also experimented with such approaches; however in our experience the highly non-linear mapping between joints and 2d points makes learning and inference harder and more computationally expensive. Consequently, we opted for estimating 3d joints directly.

\section{Solution methodology}

Our goal is to estimate body joint locations in 3-dimensional space given a 2-dimensional input. Formally, our input is a series of 2d points $\mathbf{x} \in \mathbb{R}^{2n}$, and our output is a series of points in 3d space $\mathbf{y} \in \mathbb{R}^{3n}$. We aim to learn a function $f^*: \mathbb{R}^{2n} \rightarrow \mathbb{R}^{3n}$ that minimizes the prediction error over a dataset of $N$ poses:
\begin{equation}
  f^* = \min_f \frac{1}{N} \sum_{i=1}^{N} \mathcal{L} \left( f(\mathbf{x}_i) -  \mathbf{y}_i \right).
\end{equation}

In practice, $\mathbf{x}_i$ may be obtained as ground truth 2d joint locations under known camera parameters, or using a 2d joint detector. It is also common to predict the 3d positions relative to a fixed global space with respect to its root joint, resulting in a slightly lower-dimensional output.

We focus on systems where $f^*$ is a deep neural network, and strive to find a simple, scalable and efficient architecture that performs well on this task. These goals are the main rationale behind the design choices of our network.

\subsection{Our approach -- network design}

Figure~\ref{fig:diag} shows a diagram with the basic building blocks of our architecture.
Our approach is based on a simple, deep, multilayer neural network with batch normalization~\cite{batch-norm}, dropout~\cite{dropout} and Rectified Linear Units (RELUs)~\cite{relu}, as well as residual connections~\cite{he2016deep}.
Not depicted are two extra linear layers: one applied directly to the input, which increases its dimensionality to $1024$, and one applied before the final prediction, that produces outputs of size $3n$.
In most of our experiments we use 2 residual blocks, which means that we have 6 linear layers in total, and our model contains between 4 and 5 million trainable parameters.

Our architecture benefits from multiple relatively recent improvements on the optimization of deep neural networks, which have mostly appeared in the context of very deep convolutional neural networks and have been the key ingredient of state-of-the-art systems submitted to the ILSVRC (Imagenet~\cite{imagenet}) benchmark. As we demonstrate, these contributions can also be used to improve generalization on our 2d-to-3d pose estimation task.

\paragraph{2d/3d positions} Our first design choice is to use 2d and 3d points as inputs and outputs, in contrast to recent work that has used raw images~\cite{volumetric,tekin2016structured,zhou2016deep,park20163d,tekin2016direct,ghezelghieh2016learning,du2016marker,li2015maximum,zhou2016sparseness} or 2d probability distributions~\cite{zhou2016sparseness,volumetric} as inputs, and 3d probabilities~\cite{volumetric}, 3d motion parameters~\cite{zhou2016deep} or basis pose coefficients and camera parameter estimation~\cite{Ramakrishna:2012,akhter-and-black,keep-it-simpl,zhou2016sparseness,zhou2016sparse} as outputs. While 2d detections carry less information, their low dimensionality makes them very appealing to work with; for example, one can easily store the entire Human3.6M dataset in the GPU while training the network, which reduces overall training time, and considerably allowed us to accelerate the search for network design and training hyperparameters. 

\paragraph{Linear-RELU layers} Most deep learning approaches to 3d human pose estimation are based on convolutional neural networks, which learn translation-invariant filters that can be applied to entire images~\cite{tekin2016structured,volumetric,park20163d,ghezelghieh2016learning,li2015maximum}, or 2-dimensional joint-location heatmaps~\cite{volumetric, zhou2016sparseness}. However, since we are dealing with low-dimensional points as inputs and outputs, we can use simpler and less computationally expensive linear layers. RELUs~\cite{relu} are a standard choice to add non-linearities in deep neural networks.

\paragraph{Residual connections} We found that residual connections, recently proposed as a technique to facilitate the training of very deep convolutional neural networks~\cite{he2016deep}, improve generalization performance and reduce training time. In our case, they helped us reduce error by about 10\%.

\paragraph{Batch normalization and dropout}
While a simple network with the three components described above achieves good performance on 2d-to-3d pose estimation when trained on ground truth 2d positions, we have discovered that it does not perform well when trained on the output of a 2d detector, or when trained on 2d ground truth and tested on noisy 2d observations. Batch normalization~\cite{batch-norm} and dropout~\cite{dropout} improve the performance of our system in these two cases, while resulting in a slight increase of train- and test-time.

\paragraph{Max-norm constraint}
We also applied a constraint on the weights of each layer so that their maximum norm is less than or equal to 1.
Coupled with batch normalization, we found that this stabilizes training and improves generalization when the distribution differs between training and test examples.

\subsection{Data preprocessing}

We apply standard normalization to the 2d inputs and 3d outputs by subtracting the mean and dividing by the standard deviation. 
Since we do not predict the global position of the 3d prediction, we zero-centre the 3d poses around the hip joint (in line with previous work and the standard protocol of Human3.6M).

\paragraph{Camera coordinates}
In our opinion, it is unrealistic to expect an algorithm to infer the 3d joint positions in an arbitrary coordinate space, given that any translation or rotation of such space would result in no change in the input data.
A natural choice of global coordinate frame is the camera frame~\cite{tekin2016direct,zhou2016sparseness,li2015maximum,volumetric,du2016marker,zhou2016deep} since this makes the 2d to 3d problem similar across different cameras, implicitly enabling more training data per camera and preventing overfitting to a particular global coordinate frame.
We do this by rotating and translating the 3d ground-truth according to the inverse transform of the camera.
A direct effect of inferring 3d pose in an arbitrary \emph{global} coordinate frame is the failure to regress the global orientation of the person, which results in large errors in all joints. Note that the definition of this coordinate frame is arbitrary and does not mean that we are exploiting pose ground truth in our tests.

\paragraph{2d detections} 
We obtain 2d detections using the state-of-the-art stacked hourglass network of Newell~\etal~\cite{stacked-hourglass}, pre-trained on the MPII dataset~\cite{mpii}. Similar to previous work \cite{tekin2016direct, distance-matrix,li2015maximum,h36m,park20163d}, we use the bounding boxes provided with H3.6M to estimate the centre of the person in the image.
We crop a square of size $440 \times 440$ pixels around this computed centre to the detector (which is then resized to $256 \times 256$ by stacked hourglass).
The average error between these detections and the ground truth 2d landmarks is 15 pixels, which is slightly higher than the 10 pixels reported by Moreno-Noguer~\cite{distance-matrix} using CPM~\cite{cpm} on the same dataset.
We prefer stacked hourglass over CPM because (a) it has shown slightly better results on the MPII dataset, and (b) it is about 10 times faster to evaluate, which allowed us to compute detections over the entire H3.6M dataset.

\begin{table}
\centering
\begin{tabular}{@{}lrrr@{}}
\toprule
 & DMR~\cite{distance-matrix} & Ours & $\Delta$\\
\midrule
GT/GT                       & 62.17 & 37.10 & 25.07\\
GT/GT + $\mathcal{N}(0,5)$  & 67.11  & 46.65 & 20.46\\
GT/GT + $\mathcal{N}(0,10)$ & 79.12 & 52.84 & 26.28\\
GT/GT + $\mathcal{N}(0,15)$ & 96.08 & 59.97 & 36.11\\
GT/GT + $\mathcal{N}(0,20)$ & 115.55 & 70.24 & 45.31\\
\midrule
GT/CPM~\cite{cpm} & 76.47 & -- & -- \\
GT/SH~\cite{stacked-hourglass}  & -- & 60.52 & --\\
\bottomrule
\end{tabular}
\vspace{3mm}
\caption{Performance of our system on Human3.6M under protocol \#2. (Top) Training and testing on ground truth 2d joint locations plus different levels of additive gaussian noise. (Bottom) Training on ground truth and testing on the output of a 2d detector.}
\label{tab:gt}
\end{table}

We have also fine-tuned the stacked hourglass model on the Human3.6M dataset (originally pre-trained on MPII),
which obtains more accurate 2d joint detections on our target dataset and further reduces the 3d pose estimation error. We used all the default parameters of stacked hourglass, except for minibatch size which we reduced from 6 to 3 due to memory limitations on our GPU. We set the learning rate to $2.5 \times 10^{-4}$, and train for 40\,000 iterations.

\begin{table*}
\centering
\footnotesize
\hspace{-3mm}
\tabcolsep=0.75mm
\begin{tabular}{@{}lrrrrrrrrrrrrrrrr@{}}
\toprule
Protocol \#1 & Direct. & Discuss & Eating & Greet & Phone & Photo & Pose & Purch. & Sitting & SitingD & Smoke & Wait & WalkD & Walk & WalkT & Avg\\
\midrule
LinKDE ~\cite{h36m} (SA)  & 132.7 & 183.6 & 132.3 & 164.4 & 162.1 & 205.9 & 150.6 & 171.3 & 151.6 & 243.0 & 162.1 & 170.7 & 177.1 & 96.6 & 127.9 & 162.1\\
Li~\etal~\cite{li2015maximum} (MA) & -- & 136.9 & 96.9 & 124.7 & -- & 168.7 & -- & -- & -- & -- & -- & -- & 132.2 & 70.0 & -- & --\\
Tekin~\etal~\cite{tekin2016direct} (SA) & 102.4 & 147.2 & 88.8 & 125.3 & 118.0 & 182.7 & 112.4 & 129.2 & 138.9 & 224.9 & 118.4 & 138.8 & 126.3 & 55.1 & 65.8 & 125.0\\
Zhou~\etal~\cite{zhou2016sparseness} (MA) & 87.4 & 109.3 & 87.1 & 103.2 & 116.2 & 143.3 & 106.9 & 99.8 & 124.5 & 199.2 & 107.4 & 118.1 & 114.2 & 79.4 & 97.7 & 113.0\\
Tekin~\etal~\cite{tekin2016structured} (SA) & -- & 129.1 & 91.4 & 121.7 & -- & 162.2 & -- & -- & -- & -- & -- & -- & 130.5 & 65.8 & -- & --\\
Ghezelghieh~\etal~\cite{ghezelghieh2016learning} (SA) & 80.3 & 80.4 & 78.1 & 89.7 & -- & -- & -- & -- & -- & -- & -- & -- & -- & 95.1 & 82.2 & --\\
Du~\etal~\cite{du2016marker} (SA) & 85.1 & 112.7 & 104.9 & 122.1 & 139.1 & 135.9 & 105.9 & 166.2 & 117.5 & 226.9 & 120.0 & 117.7 & 137.4 & 99.3 & 106.5 & 126.5\\
Park~\etal~\cite{park20163d} (SA) & 100.3 & 116.2 & 90.0 & 116.5 & 115.3 & 149.5 & 117.6 & 106.9 &  137.2 & 190.8 & 105.8 & 125.1 & 131.9 & 62.6 & 96.2 & 117.3\\
Zhou~\etal~\cite{zhou2016deep} (MA)   & 91.8 & 102.4 & 96.7 & 98.8 & 113.4 & 125.2 & 90.0 & 93.8 & 132.2 & 159.0 & 107.0 & 94.4 & 126.0 & 79.0 & 99.0 & 107.3\\
Pavlakos~\etal~\cite{volumetric} (MA) & 67.4 & 71.9 & 66.7 & 69.1 & 72.0 & \bf{77.0} & 65.0 & 68.3 & 83.7 & 96.5 & 71.7 & 65.8 & 74.9 & 59.1 & 63.2 & 71.9\\
\midrule
Ours (SH detections) (SA) & 61.6 & 73.4 & 63.3 & 58.3 & 91.8 & 93.6 & 66.3 & 62.0 & 91.7 & 109.4 & 75.7 & 86.5 & 67.2 & 51.2 & 52.3 & 73.6\\
Ours (SH detections) (MA) & \underline{53.3} & \underline{60.8} & \underline{62.9} & \underline{62.7} & \underline{86.4} & 82.4 & \underline{57.8} & \underline{58.7} & \underline{81.9} & \underline{99.8} & \underline{69.1} & \underline{63.9} & \underline{67.1} & \underline{50.9} & \underline{54.8} & \underline{67.5} \\
Ours (SH detections FT) (MA) & \bf{51.8}&  \bf{56.2}&	\bf{58.1}&	\bf{59.0}&	\bf{69.5}&	\underline{78.4}&	\bf{55.2}&	\bf{58.1}&	\bf{74.0}&	\bf{94.6}&	\bf{62.3}&	\bf{59.1}&	\bf{65.1}&	\bf{49.5}&	\bf{52.4}&	\bf{62.9}\\
\midrule
Ours (GT detections) (MA) & 37.7& 	44.4& 	40.3& 	42.1& 	48.2& 	54.9& 	44.4& 	42.1& 	54.6& 	58.0& 	45.1& 	46.4& 	47.6& 	36.4& 	40.4& 	45.5\\
\bottomrule
\end{tabular}
\vspace{3mm}
\caption{Detailed results on Human3.6M~\cite{h36m} under Protocol \#1 (no rigid alignment in post-processing). SH indicates that we trained and tested our model with Stacked Hourglass~\cite{stacked-hourglass} detections as input, and FT indicates that the 2d detector model was fine-tuned on H3.6M. GT detections denotes that the groundtruth 2d locations were used. SA indicates that a model was trained for each action, and MA indicates that a single model was trained for all actions.}
\label{tab:h36m}
\end{table*}

\paragraph{Training details}
We train our network
for 200 epochs
using Adam~\cite{adam}, a starting learning rate of 0.001 and exponential decay, using mini-batches of size 64. Initially, the weights of our linear layers are set using Kaiming initialization~\cite{he2015delving}. We implemented our code using Tensorflow,
which takes around 5ms for a forward+backward pass, and around 2ms for a forward pass on a 
Titan Xp
GPU. 
This means that, coupled with a state-of-the-art realtime 2d detector (\eg, ~\cite{cpm}), our network could be part of full pixels-to-3d system that runs in real time.

One epoch of training on the entire Human3.6M dataset can be done in around 2 minutes, which allowed us to extensively experiment with multiple variations of our architecture and training hyperparameters.

\begin{table*}
\centering
\footnotesize
\hspace{-3mm}
\tabcolsep=0.6mm
\begin{tabular}{@{}lrrrrrrrrrrrrrrrr@{}}
\toprule
Protocol \#2 & Direct. & Discuss & Eating & Greet & Phone & Photo & Pose & Purch. & Sitting & SitingD & Smoke & Wait & WalkD & Walk & WalkT & Avg\\
\midrule
Akhter \& Black~\cite{akhter-and-black}* (MA) 14j & 199.2 & 177.6 & 161.8 & 197.8 & 176.2 & 186.5 & 195.4 & 167.3 & 160.7 & 173.7 & 177.8 & 181.9 & 176.2 & 198.6 & 192.7 & 181.1\\
Ramakrishna~\etal~\cite{Ramakrishna:2012}* (MA) 14j & 137.4 & 149.3 & 141.6 & 154.3 & 157.7 & 158.9 & 141.8 & 158.1 & 168.6 & 175.6 & 160.4 & 161.7 & 150.0 & 174.8 & 150.2 & 157.3\\
Zhou~\etal~\cite{zhou2016sparse}* (MA) 14j & 99.7 & 95.8 & 87.9 & 116.8 & 108.3 & 107.3 & 93.5 & 95.3 & 109.1 & 137.5 & 106.0 & 102.2 & 106.5 & 110.4 & 115.2 & 106.7\\
Bogo~\etal~\cite{keep-it-simpl} (MA) 14j & 62.0 & 60.2 & 67.8 & 76.5 & 92.1 & 77.0 & 73.0 & 75.3 & 100.3 & 137.3 & 83.4 & 77.3 & 86.8 & 79.7 & 87.7 & 82.3\\
Moreno-Noguer~\cite{distance-matrix} (MA) 14j & 66.1 & 61.7 & 84.5 & 73.7 & 65.2 & 67.2 & 60.9 & 67.3 & 103.5 & 74.6 & 92.6 & 69.6 & 71.5 & 78.0 & 73.2 & 74.0\\
Pavlakos~\etal~\cite{volumetric} (MA) 17j & -- & -- & -- & -- & -- & -- & -- & -- & -- & -- & -- & -- & -- & -- & -- & \underline{51.9}\\
\midrule
Ours (SH detections) (SA) 17j & 50.1 &  59.5 & 51.3 & 56.9 & 68.5 & 67.5 & 51.0 & 47.2 & 68.5 & 85.6 & 61.2 & 67.0 & 55.1 & 41.1 & 45.5 & 58.5  \\
Ours (SH detections) (MA) 17j & \underline{42.2}&	\underline{48.0}&	\underline{49.8}&	\underline{50.8}&	\underline{61.7}&	\underline{60.7}&	\underline{44.2}&	\underline{43.6}&	\underline{64.3}&	\underline{76.5}&	\underline{55.8}&	\underline{49.1}&	\underline{53.6}&	\underline{40.8}&	\underline{46.4}&	52.5\\
Ours (SH detections FT) (MA) 17j & \bf{39.5} & \bf{43.2}&	\bf{46.4}&	\bf{47.0}&	\bf{51.0}&	\bf{56.0}&	\bf{41.4}&	\bf{40.6}&	\bf{56.5}&	\bf{69.4}&	\bf{49.2}&	\bf{45.0}&	\bf{49.5}&	\bf{38.0}&	\bf{43.1}&	\bf{47.7}\\
\midrule
Ours (SH detections) (SA) 14j & 44.8 & 52.0 & 44.4 & 50.5 &	61.7 & 59.4 & 45.1 & 41.9 &	66.3 & 77.6 & 54.0 & 58.8 & 49.0 & 35.9 & 40.7 &  52.1\\
\bottomrule
\end{tabular}

\vspace{3mm}
\caption{Detailed results on Human3.6M~\cite{h36m} under protocol \#2 (rigid alignment in post-processing).
The 14j (17j) annotation indicates that the body model considers 14 (17) body joints.
The results of all approaches are obtained from the original
papers, except for (*), which were obtained from~\cite{keep-it-simpl}.}
\label{tab:h36m_2}
\end{table*}

\section{Experimental evaluation}

\paragraph{Datasets and protocols} We focus our numerical evaluation on two standard datasets for 3d human pose estimation: HumanEva~\cite{heva} and Human3.6M~\cite{h36m}. We also show qualitative results on the MPII dataset~\cite{mpii}, for which the ground truth 3d is not available. 

Human3.6M is, to the best of our knowledge, currently the largest publicly available datasets for human 3d pose estimation. The dataset consists of 3.6 million images featuring 7 professional actors performing 15 everyday activities such as walking, eating, sitting, making a phone call and engaging in a discussion. 2d joint locations and 3d ground truth positions are available, as well as projection (camera) parameters and body proportions for all the actors. HumanEva, on the other hand, is a smaller dataset that has been largely used to benchmark previous work over the last decade. MPII is a standard dataset for 2d human pose estimation based on thousands of short youtube videos.

On Human3.6M we follow the standard protocol, using subjects 1, 5, 6, 7, and 8 for training, and subjects 9 and 11 for evaluation. We report the average error in millimetres between the ground truth and our prediction across all joints and cameras, after alignment of the root (central hip) joint. Typically, training and testing is carried out independently in each action. We refer to this as protocol \#1.
However, in some of our baselines, the prediction has been further aligned with the ground truth via a rigid transformation (\eg \cite{keep-it-simpl, distance-matrix}). We call this post-processing protocol \#2. Similarly, some recent methods have trained one model for all the actions, as opposed to building action-specific models. We have found that this practice consistently improves results, so we report results for our method under these two variations.
In HumanEva, training and testing is done on all subjects and in each action separately, and the error is always computed after a rigid transformation.

\subsection{Quantitative results}

\paragraph{An upper bound on 2d-to-3d regression}
Our method, based on direct regression from 2d joint locations, naturally depends on the quality of the output of a 2d pose detector, and achieves its best performance when it uses ground-truth 2d joint locations.

We followed Moreno-Noguer~\cite{distance-matrix} and tested under different levels of Gaussian noise a system originally trained with 2d ground truth. The results can be found in Table~\ref{tab:gt}. Our method largely outperforms the Distance-Matrix method~\cite{distance-matrix} for all levels of noise, and achieves a peak performance of $37.10$ mm of error when it is trained on ground truth 2d projections. This is about 43\% better than the best result we are aware of reported on ground truth 2d joints~\cite{distance-matrix}.
Moreover, note that this result is also about 30\% better than the $51.9$ mm reported by Pavlakos~\etal~\cite{volumetric}, which is the best result on Human3.6M that we aware of -- however, their result does not use ground truth 2d locations, which makes this comparison unfair.

Although every frame is evaluated independently, and we make no use of time, we note that the predictions produced by our network are quite smooth.
A video with these and more qualitative results can be found at \url{https://youtu.be/Hmi3Pd9x1BE}.

\paragraph{Robustness to detector noise}

To further analyze the robustness of our approach, we also experimented with testing the system (always trained with  ground truth 2d locations) with (noisy) 2d detections from images.
These results are also reported at the bottom of Table~\ref{tab:gt}.\footnote{This was, in fact, the protocol used in the main result of~\cite{distance-matrix}.} In this case, we also outperform previous work, and demonstrate that our network can perform reasonably well when trained on ground truth and tested on the output of a 2d detector.

\begin{table}
\centering
\small
\hspace{-3mm}
\tabcolsep=1.0mm
\begin{tabular}{@{}l |lll |lll |l @{}}
\toprule
& \multicolumn{3}{c}{Walking} & \multicolumn{3}{c}{Jogging} &\\
& S1 & S3 & S3 & S1 & S2 & S3 & Avg\\
\midrule
Radwan~\etal~\cite{radwan2013monocular}   & 75.1 & 99.8 & 93.8 & 79.2 & 89.8 & 99.4 & 89.5\\
Wang~\etal~\cite{wang2014robust}          & 71.9 & 75.7 & 85.3 & 62.6 & 77.7 & 54.4 & 71.3\\
Simo-Serra~\etal~\cite{simo2013joint}     & 65.1 & 48.6 & 73.5 & 74.2 & 46.6 & 32.2 & 56.7\\
Bo~\etal~\cite{bo2010twin}                & 46.4 & 30.3 & 64.9 & 64.5 & 48.0 & 38.2 & 48.7\\
Kostrikov~\etal~\cite{kostrikov2014depth} & 44.0 & 30.9 & 41.7 & 57.2 & 35.0 & 33.3 & 40.3\\
Yasin~\etal~\cite{yasin2016dual}          & 35.8 & 32.4 & 41.6 & 46.6 & 41.4 & 35.4 & 38.9\\
Moreno-Noguer~\cite{distance-matrix}      & {\bf19.7} & {\bf13.0} & {\bf24.9} & 39.7 & 20.0 & 21.0 & 26.9\\
Pavlakos~\etal~\cite{volumetric}          & 22.1 & 21.9 & \underline{29.0} & 29.8 & 23.6 & 26.0 & 25.5\\
Ours (SH detections)                      & {\bf19.7} & \underline{17.4} & 46.8 & {\bf26.9} & {\bf18.2} & {\bf18.6} & {\bf24.6}\\
\bottomrule
\end{tabular}
\vspace{3mm}
\caption{Results on the HumanEva~\cite{heva} dataset, and comparison with previous work.
}
\label{tab:heva}
\end{table}

\paragraph{Training on 2d detections}
While using 2d ground truth at train and test time is interesting to characterize the performance of our network, in a practical application our system has to work with the output of a 2d detector. We report our results on protocol \#1 of Human3.6M in Table~\ref{tab:h36m}. Here, our closest competitor is the recent volumetric prediction method of Pavlakos~\etal~\cite{volumetric}, which uses a stacked-hourglass architecture, is trained end-to-end on Human3.6M, and uses a single model for all actions.
Our method outperforms this state-of-the-art result by $4.4$ mm even when using out-of-the-box stacked-hourglass detections, and more than doubles the gap to $9.0$ mm when the 2d detector is fine-tuned on H3.6M. Our method also consistently outperforms previous work in all but one of the 15 actions of H3.6M.

Our results on Human3.6M under protocol \#2 (using a rigid alignment with the ground truth), are shown in Table~\ref{tab:h36m_2}. 
Although our method is slightly worse than previous work with out-of-the-box detections, it comes first when we use fine-tuned detections.

Finally, we report results on the HumanEva dataset in Table~\ref{tab:heva}. In this case, we obtain the best result to date in 3 out of 6 cases, and overall the best average error for actions \textit{Jogging} and \textit{Walking}. Since this dataset is rather small, and the same subjects show up on the train and test set, we do not consider these results to be as significant as those obtained by our method in Human3.6M.

\begin{figure*}
  \includegraphics[width=\linewidth,trim=70mm 66mm 60mm 25mm,clip=true]{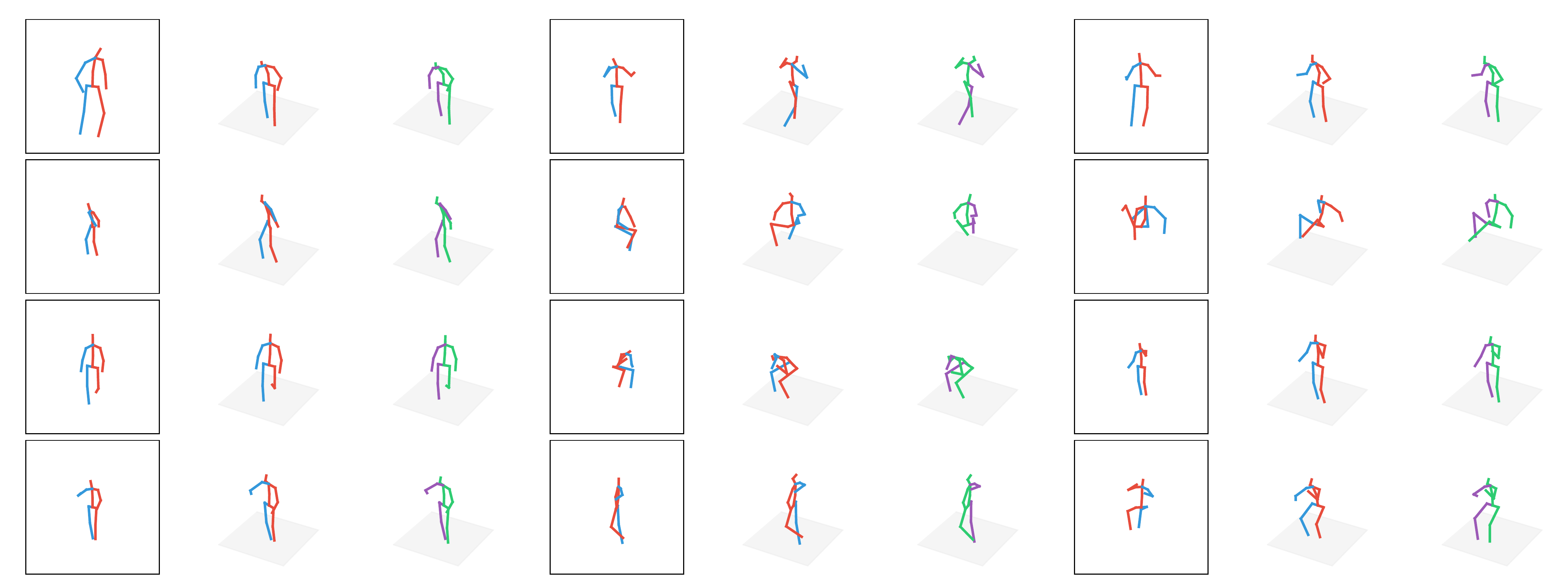}
  \vspace{-3mm}
  \caption{Example output on the test set of Human3.6M. Left: 2d observation. Middle: 3d ground truth. Right (green): our 3d predictions.
  }
  \label{fig:qualitative}
\end{figure*}

\paragraph{Ablative and hyperparameter analysis}

We also performed an ablative analysis to better understand the impact of the design choices of our network. Taking as a basis our non-fine tuned MA model, we present those results in Table~\ref{tab:ablative}. Removing dropout or batch normalization leads to $3$-to$8$ mm of increase in error, and residual connections account for a gain of about $8$ mm in our result.
However, not pre-processing the data to the network in camera coordinates results in error above $100$ mm -- substantially worse than state-of-the-art performance.

Last but not least, we analyzed the sensitivity of our network to depth and width. Using a single residual block results in a loss of $6$ mm, and performance is saturated after 2 blocks. Empirically, we observed that decreasing the layers to $512$ dimensions gave worse performance, while layers with $2\,048$ units were much slower and did not seem to increase the accuracy.

\subsection{Qualitative results}

\begin{figure*}[t]
  \includegraphics[width=\linewidth,trim=70mm 25mm 63mm 69mm,clip=true]{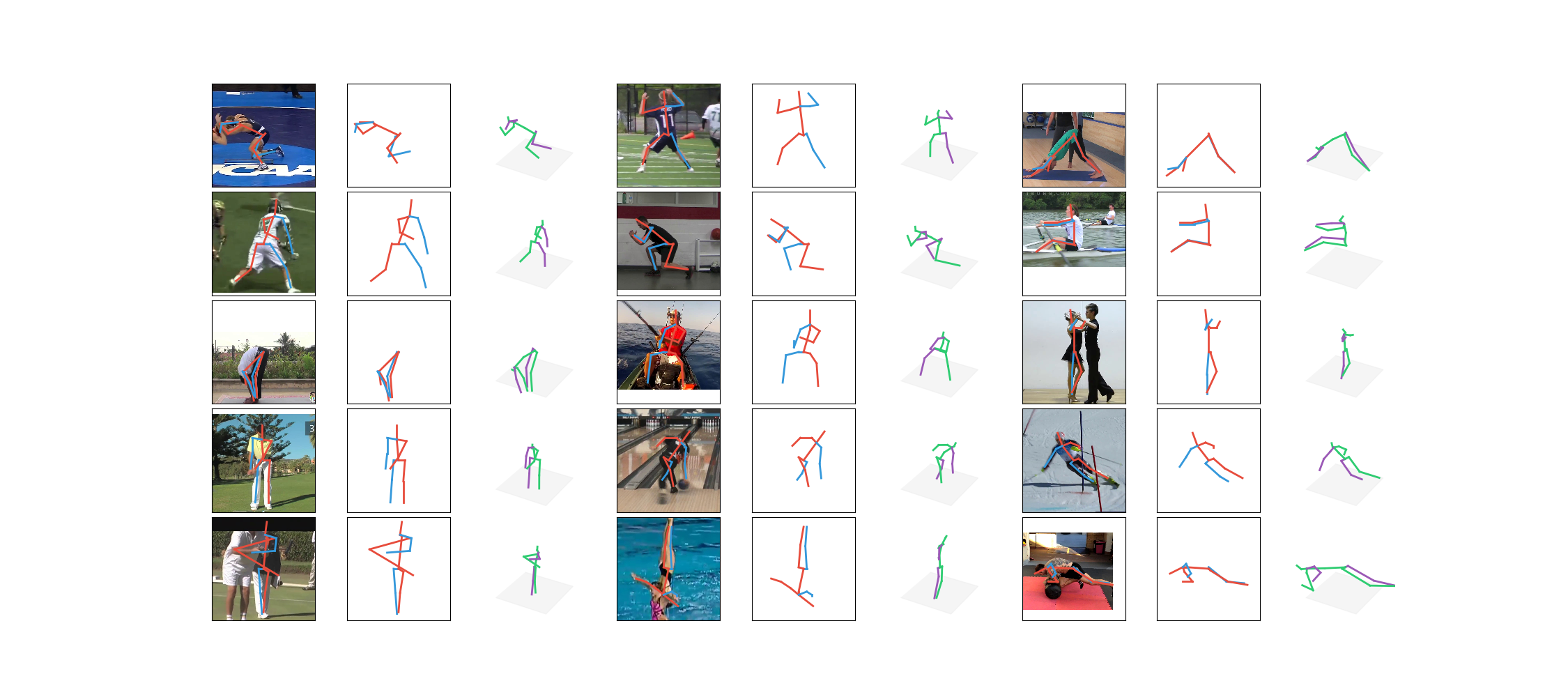}
  \vspace{-3mm}
  \caption{Qualitative results on the MPII test set. Observed image, 2d detection with Stacked Hourglass~\cite{stacked-hourglass}, (in green) our 3d prediction. The bottom 3 examples are typical failure cases, where either the 2d detector has failed badly (left), or slightly (right). In the middle, the 2d detector does a fine job, but the person is upside-down and Human3.6M does not provide any similar examples -- the network still seems to predict an average pose.}
  \label{fig:mpii}
\end{figure*}

Finally, we show some qualitative results on Human3.6M in Figure~\ref{fig:qualitative}, and from images ``in the wild'' from the test set of MPII in Figure~\ref{fig:mpii}. Our results on MPII reveal some of the limitations of our approach; for example, our system cannot recover from a failed detector output, and it has a hard time dealing with poses that are not similar to any examples in H3.6M (\eg people upside-down).
Finally, in the wild most images of people do not feature full bodies, but are cropped to some extent. Our system, trained on full body poses, is currently unable to deal with such cases.

\begin{table}
\centering
\hspace{-3mm}
\begin{tabular}{@{}lrr@{}}
\toprule
 & error (mm) & $\Delta$\\
 \midrule
Ours & 67.5 & --\\
w/o batch norm & 88.5 & 21.0\\
w/o dropout & 71.4 & 3.9\\
w/o batch norm w/o dropout & 76.0 & 8.5\\
w/o residual connections&  75.8 & 8.3\\
w/o camera coordinates & 101.1 & 33.6\\
\midrule
1 block & 74.2 & 6.7\\
2 blocks (Ours) & 67.5 & --\\
4 blocks & 69.3 & 1.8\\
8 blocks & 69.7 & 2.4\\
\bottomrule
\end{tabular}
\vspace{3mm}
\caption{Ablative and hyperparameter sensitivity analysis.}
\label{tab:ablative}
\end{table}

\section{Discussion}

Looking at Table~\ref{tab:h36m}, we see a generalized increase in error when training with SH detections as opposed to training with ground truth 2d across all actions -- as one may well expect. There is, however, a particularly large increase in the classes \emph{taking photo}, \emph{talking on the phone}, \emph{sitting} and \emph{sitting down}. We hypothesize that this is due to the severe self-occlusions in these actions -- for example, in some \emph{phone} sequences, we never get to see one of the hands of the actor. Similarly, in \emph{sitting} and \emph{sitting down}, the legs are often aligned with the camera viewpoint, which results in large amounts of foreshortening.

\paragraph{Further improvements}
The simplicity of our system suggests multiple directions of improvement in future work.
For example, we note that stacked hourglass produces final joint detection heatmaps of size $64 \times 64$, and thus a larger output resolution might result in more fine-grained detections, moving our system closer to its performance when trained on ground truth.
Another interesting direction is to use multiple samples from the 2d stacked hourglass heatmaps to estimate an expected gradient -- \emph{{\`a} la} policy gradients, commonly used in reinforcement learning -- so as to train a network end-to-end.
Yet another idea is to emulate the output of 2d detectors using 3-dimensional mocap databases and ``fake'' camera parameters for data augmentation, perhaps following the adversarial approach of Shrivastava~\etal~\cite{shrivastava2016learning}.
Learning to estimate coherently the depth of each person in the scene is an interesting research path, since it would allow our system to work on 3d pose estimation of multiple people.
Finally, our architecture is simple, and it is likely that further research into network design could lead to better results on 2d-to-3d systems.

\subsection{Implications of our results}
We have demonstrated that a relatively simple deep feedforward neural network
can achieve a remarkably low error rate on 3d human pose estimation. Coupled with a state-of-the-art 2d detector, our system 
obtains the best results on 3d pose estimation to date.

Our results stand in contrast to recent work, which has focused on deep, end-to-end systems trained from pixels to 3d positions, and contradicts the underlying hypothesis that justify the complexity of recent state-of-the-art approached to 3d human pose estimation.
For example, the volumetric regression approach of~\cite{volumetric}~\etal is based on the hypothesis that directly regressing 3d points is inherently difficult, and regression in a volumetric space would provide easier gradients for the network (see Table 1 in~\cite{volumetric}).
Although we agree that image content should help to resolve challenging ambiguous cases (consider for example the classic turning ballerina optical illusion), competitive 3d pose estimation from 2d points can be achieved with simple high capacity systems. This might be related to the latent information about subtle body and motion traits existing in 2d joint stimuli, such as gender, which can be perceived by people~\cite{troje2006adaptation}.
Similarly, the use of a distance matrix as a body representation in~\cite{distance-matrix} is justified by the claim that invariant, human-designed features should boost the accuracy of the system. However, our results show that well trained systems can outperform these particular features in a simple manner.
It would be interesting to see whether a combination of joint distances and joint positions boost the performance even further -- we leave this for future work.

\section{Conclusions and future work}

We have shown that a simple, fast and lightweight deep neural network can achieve surprisingly accurate results in the task of 2d-to-3d human pose estimation; and coupled with a state-of-the-art 2d detector, our work results in an easy-to-reproduce, yet high-performant baseline that outperforms the state of the art in 3d human pose estimation.

Our accuracy in 3d pose estimation from 2d ground truth suggest that, although 2d pose estimation is considered a close to solved problem, it remains as one of the main causes for error in the 3d human pose estimation task. 
Moreover, our work represents poses in simple 2d and 3d coordinates, which suggests that finding invariant (and more complex) representations of the human body, as has been the focus of recent work, might either not be crucial, or have not been exploited to its full potential.

Finally, given its simplicity and the rapid development in the field, we like to think of our work as a future baseline, rather than a full-fledged system for 3d pose estimation. This suggests multiple directions of future work. For one, our network currently does not have access to visual evidence; we believe that adding this information to our pipeline, either via fine-tuning of the 2d detections or through multi-sensor fusion will lead to further gains in performance. On the other hand, our architecture is similar to a multi-layer perceptron, which is perhaps the simplest architecture one may think of. We believe that a further exploration of the network architectures will result in improved performance. These are all interesting areas of future work.

\paragraph{Acknowledgments} The authors thank NVIDIA for the donation of GPUs used in this research.
Julieta was supported in part by the Perceiving Systems group at the Max Planck Institute for Intelligent Systems.
This research was supported in part by the Natural Sciences and Engineering Research Council of Canada (NSERC).

{\small
\bibliographystyle{ieee}
\balance
\bibliography{egbib}

\begin{thebibliography}{10}\itemsep=-1pt

\bibitem{AgarwalT04}
A.~Agarwal and B.~Triggs.
\newblock 3{D} human pose from silhouettes by relevance vector regression.
\newblock In {\em CVPR}, 2004.

\bibitem{akhter-and-black}
I.~Akhter and M.~J. Black.
\newblock Pose-conditioned joint angle limits for {3D} human pose
  reconstruction.
\newblock In {\em CVPR}, 2015.

\bibitem{mpii}
M.~Andriluka, L.~Pishchulin, P.~Gehler, and B.~Schiele.
\newblock 2d human pose estimation: New benchmark and state of the art
  analysis.
\newblock In {\em CVPR}, 2014.

\bibitem{barron_2001}
C.~Barron and I.~A. Kakadiaris.
\newblock Estimating anthropometry and pose from a single uncalibrated image.
\newblock {\em CVIU}, 81(3):269--284, 2001.

\bibitem{bo2010twin}
L.~Bo and C.~Sminchisescu.
\newblock Twin {G}aussian processes for structured prediction.
\newblock {\em IJCV}, 87(1-2), 2010.

\bibitem{bb41073}
L.~F. Bo, C.~Sminchisescu, A.~Kanaujia, and D.~N. Metaxas.
\newblock Fast algorithms for large scale conditional 3{D} prediction.
\newblock In {\em CVPR}, pages 1--8, 2008.

\bibitem{keep-it-simpl}
F.~Bogo, A.~Kanazawa, C.~Lassner, P.~Gehler, J.~Romero, and M.~J. Black.
\newblock Keep it {SMPL}: Automatic estimation of 3d human pose and shape from
  a single image.
\newblock In {\em ECCV}, 2016.

\bibitem{bulthoff1998top}
I.~B{\"u}lthoff, H.~B{\"u}lthoff, and P.~Sinha.
\newblock Top-down influences on stereoscopic depth-perception.
\newblock {\em Nature Neuroscience}, 1(3):254--257, 1998.

\bibitem{ChenR16a}
C.-H. Chen and D.~Ramanan.
\newblock 3{D} human pose estimation = 2{D} pose estimation + matching.
\newblock In {\em CVPR}, 2017.

\bibitem{imagenet}
J.~Deng, W.~Dong, R.~Socher, L.-J. Li, K.~Li, and L.~Fei-Fei.
\newblock Imagenet: A large-scale hierarchical image database.
\newblock In {\em CVPR}, 2009.

\bibitem{du2016marker}
Y.~Du, Y.~Wong, Y.~Liu, F.~Han, Y.~Gui, Z.~Wang, M.~Kankanhalli, and W.~Geng.
\newblock Marker-less 3d human motion capture with monocular image sequence and
  height-maps.
\newblock In {\em ECCV}, 2016.

\bibitem{fergus_2D_3D_iccv2015}
D.~Eigen and R.~Fergus.
\newblock Predicting depth, surface normals and semantic labels with a common
  multi-scale convolutional architecture.
\newblock In {\em ICCV}, 2015.

\bibitem{ghezelghieh2016learning}
M.~F. Ghezelghieh, R.~Kasturi, and S.~Sarkar.
\newblock Learning camera viewpoint using cnn to improve 3d body pose
  estimation.
\newblock In {\em {3DV}}, 2016.

\bibitem{gupta20143dpose}
A.~Gupta, J.~Martinez, J.~J. Little, and R.~J. Woodham.
\newblock {3D Pose from Motion for Cross-view Action Recognition via Non-linear
  Circulant Temporal Encoding}.
\newblock In {\em CVPR}, 2014.

\bibitem{he2015delving}
K.~He, X.~Zhang, S.~Ren, and J.~Sun.
\newblock Delving deep into rectifiers: Surpassing human-level performance on
  imagenet classification.
\newblock In {\em ICCV}, pages 1026--1034, 2015.

\bibitem{he2016deep}
K.~He, X.~Zhang, S.~Ren, and J.~Sun.
\newblock Deep residual learning for image recognition.
\newblock In {\em CVPR}, pages 770--778, 2016.

\bibitem{batch-norm}
S.~Ioffe and C.~Szegedy.
\newblock Batch normalization: Accelerating deep network training by reducing
  internal covariate shift.
\newblock In {\em ICML}, 2015.

\bibitem{ionescu_cvpr14}
C.~Ionescu, J.~Carreira, and C.~Sminchisescu.
\newblock Iterated second-order label sensitive pooling for 3d human pose
  estimation.
\newblock In {\em CVPR}, 2014.

\bibitem{h36m}
C.~Ionescu, D.~Papava, V.~Olaru, and C.~Sminchisescu.
\newblock {Human 3.6m: Large scale datasets and predictive methods for 3d human
  sensing in natural environments}.
\newblock {\em TPAMI}, 36(7), 2014.

\bibitem{jiang_10}
H.~Jiang.
\newblock 3d human pose reconstruction using millions of exemplars.
\newblock In {\em ICPR}, pages 1674--1677, Aug 2010.

\bibitem{adam}
D.~Kingma and J.~Ba.
\newblock Adam: A method for stochastic optimization.
\newblock In {\em ICLR}, 2015.

\bibitem{kostrikov2014depth}
I.~Kostrikov and J.~Gall.
\newblock Depth sweep regression forests for estimating 3d human pose from
  images.
\newblock In {\em BMVC}, 2014.

\bibitem{Chen85b}
H.~J. Lee and Z.~Chen.
\newblock Determination of 3{D} human body postures from a single view.
\newblock {\em Computer Vision, Graphics and Image Processing}, 30:148--168,
  1985.

\bibitem{li2015maximum}
S.~Li, W.~Zhang, and A.~B. Chan.
\newblock Maximum-margin structured learning with deep networks for 3d human
  pose estimation.
\newblock In {\em ICCV}, 2015.

\bibitem{lindeberg_shapefromtexture}
T.~Lindeberg and J.~Garding.
\newblock Shape from texture from a multi-scale perspective.
\newblock In {\em ICCV}, 1993.

\bibitem{liu_2d_3d_cvpr_2015}
F.~Liu, C.~Shen, and G.~Lin.
\newblock Deep convolutional neural fields for depth estimation from a single
  image.
\newblock In {\em CVPR}, pages 5162--5170. IEEE Computer Society, 2015.

\bibitem{distance-matrix}
F.~Moreno-Noguer.
\newblock 3d human pose estimation from a single image via distance matrix
  regression.
\newblock In {\em CVPR}, 2017.

\bibitem{bb93102}
G.~Mori and J.~Malik.
\newblock Recovering 3{D} human body configurations using shape contexts.
\newblock {\em TPAMI}, 28(7):1052--1062, July 2006.

\bibitem{relu}
V.~Nair and G.~E. Hinton.
\newblock Rectified linear units improve restricted {B}oltzmann machines.
\newblock In {\em ICML}, pages 807--814, 2010.

\bibitem{stacked-hourglass}
A.~Newell, K.~Yang, and J.~Deng.
\newblock Stacked hourglass networks for human pose estimation.
\newblock In {\em ECCV}, 2016.

\bibitem{ParameswaranC04}
V.~Parameswaran and R.~Chellappa.
\newblock View independent human body pose estimation from a single perspective
  image.
\newblock In {\em CVPR}, 2004.

\bibitem{park20163d}
S.~Park, J.~Hwang, and N.~Kwak.
\newblock 3d human pose estimation using convolutional neural networks with 2d
  pose information.
\newblock In {\em ECCV 2016 Workshops}, 2016.

\bibitem{volumetric}
G.~Pavlakos, X.~Zhou, K.~G. Derpanis, and K.~Daniilidis.
\newblock Coarse-to-fine volumetric prediction for single-image 3{D} human
  pose.
\newblock In {\em CVPR}, 2017.

\bibitem{popa_cvpr17}
A.~Popa, M.~Zanfir, and C.~Sminchisescu.
\newblock {Deep Multitask Architecture for Integrated 2D and 3D Human Sensing}.
\newblock In {\em CVPR}, 2017.

\bibitem{radwan2013monocular}
I.~Radwan, A.~Dhall, and R.~Goecke.
\newblock Monocular image 3d human pose estimation under self-occlusion.
\newblock In {\em ICCV}, 2013.

\bibitem{Ramakrishna:2012}
V.~Ramakrishna, T.~Kanade, and Y.~Sheikh.
\newblock {Reconstructing 3D Human Pose from 2D Image Landmarks}.
\newblock In {\em ECCV}, 2012.

\bibitem{roberts_63}
L.~G. Roberts.
\newblock Machine perception of three-dimensional solids.
\newblock TR 315, Lincoln Lab, MIT, Lexington, MA, May 1963.

\bibitem{RogezS16}
G.~Rogez and C.~Schmid.
\newblock Mocap-guided data augmentation for 3{D} pose estimation in the wild.
\newblock In {\em NIPS}, 2016.

\bibitem{saxena_2D_3D}
A.~Saxena, M.~Sun, and A.~Y. Ng.
\newblock Learning 3-{D} scene structure from a single still image.
\newblock In {\em ICCV}, 2007.

\bibitem{bb93266}
G.~Shakhnarovich, P.~A. Viola, and T.~J. Darrell.
\newblock Fast pose estimation with parameter-sensitive hashing.
\newblock In {\em ICCV}, 2003.

\bibitem{shrivastava2016learning}
A.~Shrivastava, T.~Pfister, O.~Tuzel, J.~Susskind, W.~Wang, and R.~Webb.
\newblock Learning from simulated and unsupervised images through adversarial
  training.
\newblock In {\em CVPR}, 2017.

\bibitem{heva}
L.~Sigal, A.~O. Balan, and M.~J. Black.
\newblock Humaneva: Synchronized video and motion capture dataset and baseline
  algorithm for evaluation of articulated human motion.
\newblock {\em IJCV}, 87(1-2), 2010.

\bibitem{simo2013joint}
E.~Simo-Serra, A.~Quattoni, C.~Torras, and F.~Moreno-Noguer.
\newblock A joint model for 2d and 3d pose estimation from a single image.
\newblock In {\em CVPR}, 2013.

\bibitem{dropout}
N.~Srivastava, G.~E. Hinton, A.~Krizhevsky, I.~Sutskever, and R.~Salakhutdinov.
\newblock Dropout: a simple way to prevent neural networks from overfitting.
\newblock {\em JMLR}, 15(1), 2014.

\bibitem{tekin2016structured}
B.~Tekin, I.~Katircioglu, M.~Salzmann, V.~Lepetit, and P.~Fua.
\newblock Structured prediction of 3d human pose with deep neural networks.
\newblock In {\em BMVC}, 2016.

\bibitem{tekin2016direct}
B.~Tekin, A.~Rozantsev, V.~Lepetit, and P.~Fua.
\newblock Direct prediction of 3d body poses from motion compensated sequences.
\newblock In {\em CVPR}, 2016.

\bibitem{troje2006adaptation}
N.~F. Troje, J.~Sadr, H.~Geyer, and K.~Nakayama.
\newblock Adaptation aftereffects in the perception of gender from biological
  motion.
\newblock {\em Journal of vision}, 6(8):7--7, 2006.

\bibitem{varol_2017}
G.~Varol, J.~Romero, X.~Martin, N.~Mahmood, M.~J. Black, I.~Laptev, and
  C.~Schmid.
\newblock Learning from synthetic humans.
\newblock In {\em CVPR}, 2017.

\bibitem{wang2014robust}
C.~Wang, Y.~Wang, Z.~Lin, A.~L. Yuille, and W.~Gao.
\newblock Robust estimation of 3d human poses from a single image.
\newblock In {\em CVPR}, 2014.

\bibitem{cpm}
S.-E. Wei, V.~Ramakrishna, T.~Kanade, and Y.~Sheikh.
\newblock Convolutional pose machines.
\newblock In {\em CVPR}, 2016.

\bibitem{WeiC09}
X.~K. Wei and J.~Chai.
\newblock Modeling 3{D} human poses from uncalibrated monocular images.
\newblock In {\em ICCV}, 2009.

\bibitem{yasin2016dual}
H.~Yasin, U.~Iqbal, B.~Kruger, A.~Weber, and J.~Gall.
\newblock A dual-source approach for 3d pose estimation from a single image.
\newblock In {\em CVPR}, 2016.

\bibitem{zhang_shapefromshading}
R.~Zhang, P.-S. Tsai, J.~E. Cryer, and M.~Shah.
\newblock Shape from shading: {A} survey.
\newblock {\em TPAMI}, 21(8):690--706, 1999.

\bibitem{zhou2016deep}
X.~Zhou, X.~Sun, W.~Zhang, S.~Liang, and Y.~Wei.
\newblock Deep kinematic pose regression.
\newblock In {\em ECCV Workshops}, 2016.

\bibitem{zhou2016sparse}
X.~Zhou, M.~Zhu, S.~Leonardos, and K.~Daniilidis.
\newblock Sparse representation for 3d shape estimation: A convex relaxation
  approach.
\newblock {\em TPAMI}, 99(1), 2016.

\bibitem{zhou2016sparseness}
X.~Zhou, M.~Zhu, S.~Leonardos, K.~G. Derpanis, and K.~Daniilidis.
\newblock Sparseness meets deepness: 3d human pose estimation from monocular
  video.
\newblock In {\em CVPR}, 2016.

\bibitem{zisserman99}
A.~Zisserman, I.~D. Reid, and A.~Criminisi.
\newblock Single view metrology.
\newblock In {\em ICCV}, 1999.

\end{thebibliography}
}

\end{document}